\documentclass[conference,9pt]{IEEEtran}
\usepackage{mathptmx} 
\usepackage{array}
\usepackage{fancyhdr}
\usepackage[normalem]{ulem}
\usepackage[hyphens]{url}
\usepackage[sort,nocompress]{cite}
\usepackage[final]{microtype}
\usepackage{tabularx}
\usepackage{cite}
\usepackage{amsmath,amssymb,amsfonts}
\usepackage{graphicx}
\usepackage{textcomp}
\usepackage{xcolor}
 
\usepackage{fancyhdr}
\usepackage[hyphens]{url}
\usepackage{pifont}
\usepackage[utf8]{inputenc} 
\usepackage[T1]{fontenc}    
\usepackage{booktabs}       
\usepackage{amsfonts}       
\usepackage{nicefrac}       
\usepackage{microtype}      
\usepackage{mathtools}
\usepackage{tikz}
\usepackage{esvect}
\usepackage{lipsum}
\usepackage{multirow}
\usepackage{multicol}
\usepackage{amsfonts}
\usepackage{caption}
\usepackage{subcaption}
\usepackage{bm}
\usepackage{cite}
\usepackage{algpseudocode}
\usepackage{stmaryrd}

\usepackage[linesnumbered]{algorithm2e}

\usepackage{multirow}
\usepackage{multicol}
\usepackage{listings}

\definecolor{codegreen}{rgb}{0,0.6,0}
\definecolor{codegray}{rgb}{0.5,0.5,0.5}
\definecolor{codepurple}{rgb}{0.58,0,0.82}
\definecolor{backcolour}{rgb}{0.95,0.95,0.92}

\lstdefinestyle{mystyle}{
    backgroundcolor=\color{white},   
    commentstyle=\color{codegreen},
    keywordstyle=\color{magenta},
    numberstyle=\tiny\color{codegray},
    stringstyle=\color{codepurple},
    basicstyle=\ttfamily\small,
    breakatwhitespace=false,         
    breaklines=true,                 
    captionpos=b,                    
    keepspaces=true,                 
    numbers=left,                    
    numbersep=5pt,                  
    showspaces=false,                
    showstringspaces=false,
    showtabs=false,                  
    tabsize=2,
}
\usepackage[bookmarks=true,breaklinks=true,letterpaper=true,colorlinks,linkcolor=black,citecolor=blue,urlcolor=black]{hyperref}

\title{Verilog-to-PyG -- A Framework for Graph Learning and Augmentation on RTL Designs}

\author{Yingjie Li$^1$, Mingju Liu$^1$, Alan Mishchenko$^2$, Cunxi Yu$^1$\\$^1$University of Maryland, College Park\\$^2$UC Berkeley}

\begin{document}
\maketitle
\pagestyle{plain}

 
\begin{abstract}
The complexity of modern hardware designs necessitates advanced methodologies for optimizing and analyzing modern digital systems. In recent times, machine learning (ML) methodologies have emerged as potent instruments for assessing design quality-of-results at the Register-Transfer Level (RTL) or Boolean level, aiming to expedite design exploration of advanced RTL configurations. In this presentation, we introduce an innovative open-source framework that translates RTL designs into graph representation foundations, which can be seamlessly integrated with the PyTorch Geometric graph learning platform. Furthermore, the Verilog-to-PyG (V2PYG) framework is compatible with the open-source Electronic Design Automation (EDA) toolchain OpenROAD, facilitating the collection of labeled datasets in an utterly open-source manner. Additionally, we will present novel RTL data augmentation methods (incorporated in our framework) that enable functional equivalent design augmentation for the construction of an extensive graph-based RTL design database. Lastly, we will showcase several using cases of V2PYG with detailed scripting examples. V2PYG can be found at \url{https://yu-maryland.github.io/Verilog-to-PyG/}.
\end{abstract}

\section{Introduction}

The increasing complexity of electronic systems has driven significant advancements in hardware design. Modern hardware designs encompass a wide range of components, from Register Transfer Level (RTL) descriptions to logic circuits. 
As the complexity grows, conventional design algorithms can suffer from their exponentially increased runtime overhead, resulting in sub-optimal solutions as the optimization search space is too large to be explored. Thus, an effective optimization and analysis technique becomes more critical. 

\begin{figure}[h]
    \centering
    \includegraphics[width=0.48\textwidth]{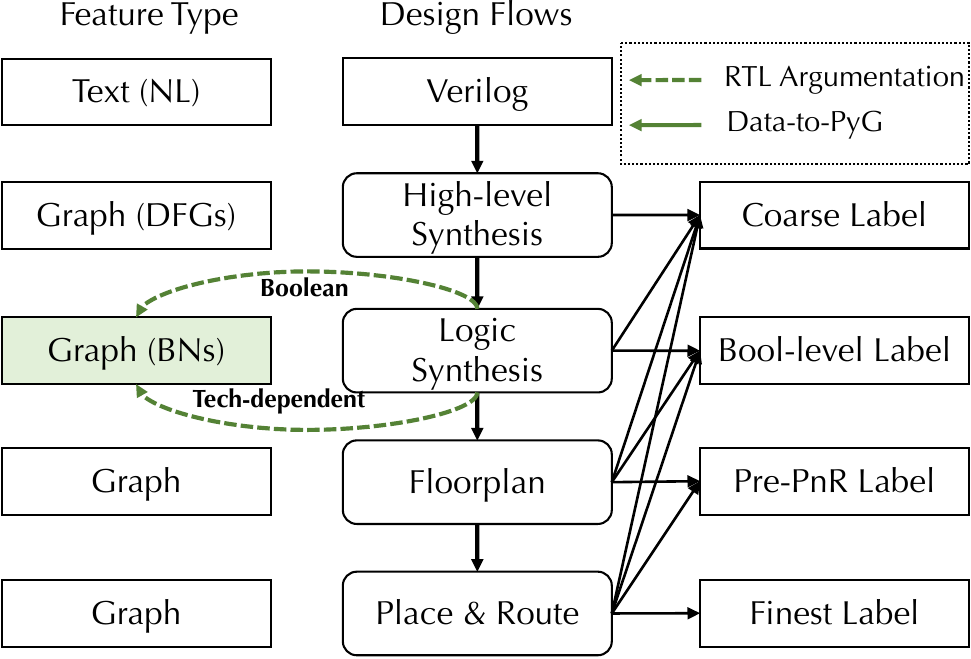}
    \caption{Overview of design features across the design flow. V2PYG currently focuses on graph data representations of RTL designs at data-flow and Boolean network level (technology independent and dependent), including coarse-to-fine data labeling via OpenROAD \cite{ajayi2019toward} and design augmentation at graph-level by incorporating computer algebra, Boolean algebra \cite{brayton2010abc}, and retiming \cite{leiserson1991retiming}.}
    \label{fig:overview}
\end{figure}

Recently, machine learning (ML), which features with its data-driven generalizability, has been applied to computer systems and electronic design automation (EDA) tasks~\cite{wu2022survey,mirhoseini2020chip,yu2018developing,yu2020flowtune,neto2022flowtune,yu2019painting,chowdhury2022bulls} as an alternative to conventional solutions. As shown in Figure \ref{fig:overview}, the optimization target in each design flow can be converted to a specific ML task. For example, for the high-level synthesis, the optimization target is the Data Flow Graphs (DFGs) \cite{ustun2020accurate,yin2023accelerating,yin2023respect,wu2021ironman,wu2022hls}, which can be formulated as graph learning problems. Similarly, for logic synthesis, the optimization target, Boolean Networks (BNs), can be formulated as graphs and applicable to graph learning. As a result, graph neural networks (GNNs) have been applied to classifying sub-circuit functionality from gate-level netlists~\cite{alrahis2021gnn, wang2022efficient, wu2023gamora}, analyzing the impacts of logic rewriting~\cite{zhao2022graph}, predicting arithmetic block boundaries~\cite{he2021graph}, and improving design exploration fidelity\cite{ustun2020accurate}.

Specifically, graph learning, particularly graph neural networks (GNNs), has emerged as a powerful method for understanding complex relationships in various domains, including social networks, biological systems, and natural language processing. In the context of hardware design, graph learning can be employed to model and analyze the intricate connections among design components, such as gates, wires, and registers, to optimize design performance and effectively analyze design characteristics. Furthermore, the embedding and features encoded within the graph, i.e., the dataset for GNNs, play a vital role in the GL performance regarding the hardware design~\cite{wu2023gamora}.

However, when it comes to applying graph learning to RTL designs, there is a notable gap in support for dataset preparation, specifically in the graph representations of these designs. Two main challenges exist: 1) The absence of a comprehensive infrastructure that links RTL designs to graph representations, aids in the construction of graph datasets, and integrates seamlessly with existing graph learning and EDA frameworks such as PyTorch Geometric (PyG) \cite{fey2019fast} and OpenROAD \cite{ajayi2019openroad}. 2) A limited range of design variations available for training and constrained search spaces for foundational RTL models, which necessitates the development of effective RTL design augmentation techniques. To tackle these challenges, we introduce an open-source infrastructure that is compatible with PyG \cite{fey2019fast} and enriches the pool of RTL design samples through functionally equivalent RTL augmentations.

In this paper, we focus on the graph data representations of RTL designs at the Boolean network level, as highlighted in the green box of Figure \ref{fig:overview}. The paper is organized as follows: Section \ref{sec:background} offers background information pertinent to graph-represented RTL designs. Section \ref{sec:framework} presents our framework, including design-specific graph generation (Section \ref{subsec:graph_extract}) and Section \ref{subsec:graph_aug} illustrates dataset augmentations for a variety of graph structures that retain the same design function. Finally, we discuss the implementation of our framework (Section \ref{subsec:abc_imp}) in ABC \cite{brayton2010abc,mishchenko2007abc}, complemented by detailed user examples (Section \ref{subsec:use_exp1}) and demonstrations of equivalence checking verification.

\section{Background}
\label{sec:background}

\subsection{Graph Learning on RTL Designs}

Since BNs and circuit netlists are naturally represented as graphs, GNNs can be leveraged to classify sub-circuit functionality from gate-level netlists~\cite{alrahis2021gnn}, predict the functionality of approximate circuits~\cite{bucher2022appgnn}, analyze impacts of circuit rewriting on functional operator detection~\cite{zhao2022graph}, and predict boundaries of arithmetic blocks~\cite{he2021graph,wang2022efficient,wang2022functionality,wu2023gamora}. Promising as they are, these approaches focus on graphs with tens of thousands of nodes and conduct training on complex designs and inference on relatively simpler ones, in which the generalization capability from simple to complex designs is not well examined. To address such challenges in learning on logic graphs, representing logic graph using universal representations have shown significant improvements in generalizability and scalability. For example, the uses of And-Inv-Graph (AIGs) has significantly boosted the performance and generaliability in contrastive graph learning \cite{zhao2022graph, li2022deepgate,shi2023deepgate2,wang2022functionality}, node classification based tasks, and reasoning \cite{wu2023gamora}. Moreover, in DFGs or similar graph representations, GNNs have also shown great potential in datapath optimization, compiler optimization, and multi-fidelity optimization such as placement and floorplanning \cite{wu2022hls,wu2022lostin,ustun2020accurate,li2022lisa, lu2020vlsi, lu2021law}.

Moreover, GNNs operate by propagating information along the edges of a given graph. Each node is initialized with a representation, which could be either a direct representation or a learnable embedding obtained from node features. Then, a GNN layer updates each node representation by integrating node representations of both itself and its neighbors in the graph. The propagation along edges extracts structural information from graphs, corresponding to structural shape hashing in conventional reasoning; after encoding Boolean functionality into node features, neighborhood aggregation is analogous to functional aggregation in conventional reasoning.
Thus, the inherent message-passing mechanism in GNNs enables simultaneous handling of structural and functional information.
This is analogous to conventional reasoning \cite{wu2023gamora,wang2022functionality}, where GNNs aggregate functional and structural approaches simultaneously.

\subsection{DAG-aware Logic Synthesis}
When applying graph learning to hardware designs, the dataset is prepared to include various graph structures even with the same functionality, which requires the framework to be able to provide graph augmentation for the dataset generation.
In logic synthesis, design variations can be generated with different logic optimization methods and different technology mapping methods. 

In logic optimization, which is conducted on the uniform functionality representation (here, we use AIG representation as a demonstration), there are mainly three methods.   

\begin{figure}[!htb]
    \centering
    \begin{subfigure}[b]{0.24\textwidth}
    \centering
        \includegraphics[width=1\textwidth]{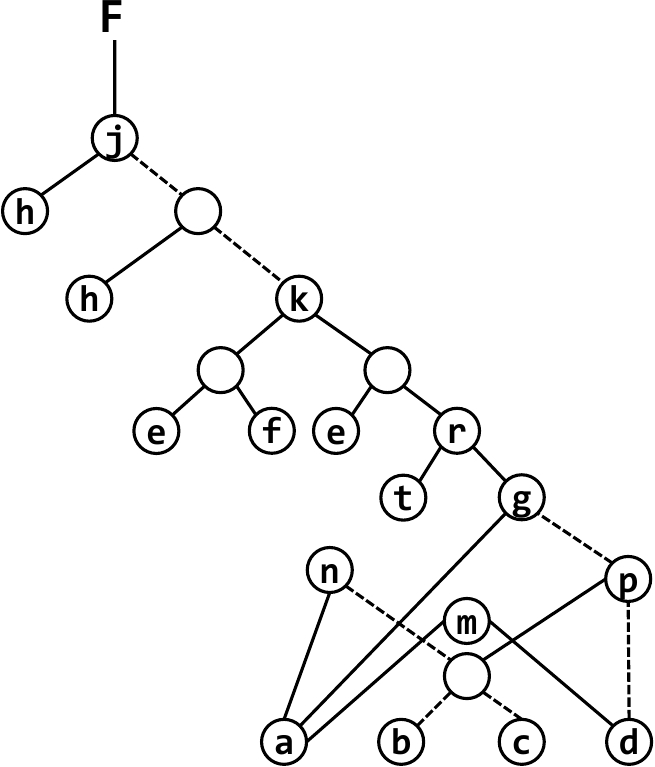}
        \caption{Original}
        \label{fig:opt_ori}
    \end{subfigure}
    \hfill
    \begin{subfigure}[b]{0.24\textwidth}
    \centering
        \includegraphics[width=1\textwidth]{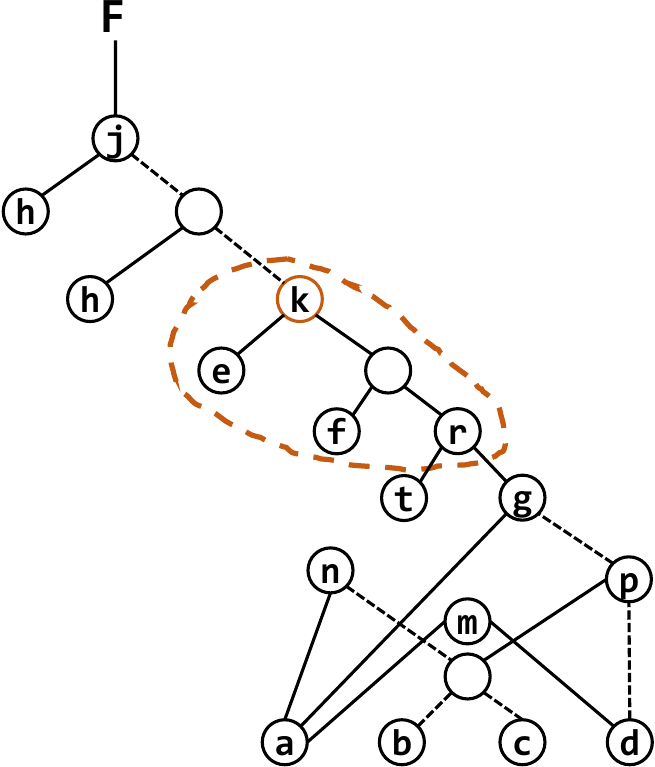}
        \caption{Rewriting}
        \label{fig:opt_rw}
    \end{subfigure}
    \hfill
    \begin{subfigure}[b]{0.24\textwidth}
    \centering
        \includegraphics[width=1\textwidth]{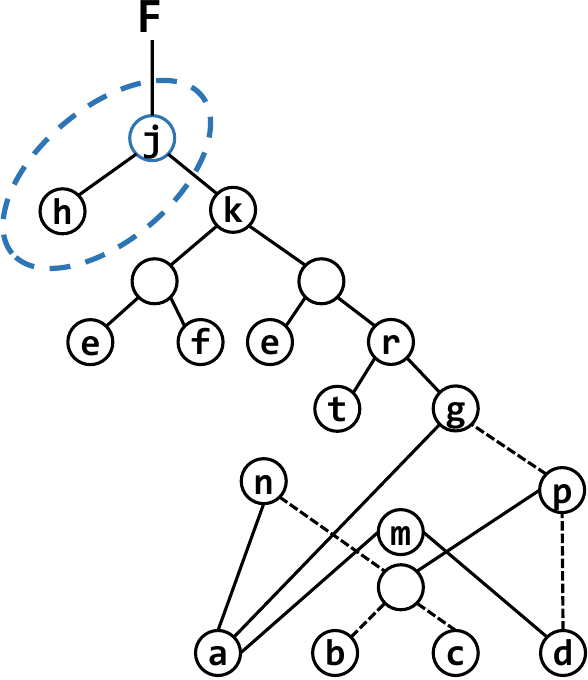}
        \caption{Refactoring}
        \label{fig:opt_rf}
    \end{subfigure}
    \hfill
    \begin{subfigure}[b]{0.24\textwidth}
    \centering
        \includegraphics[width=1\textwidth]{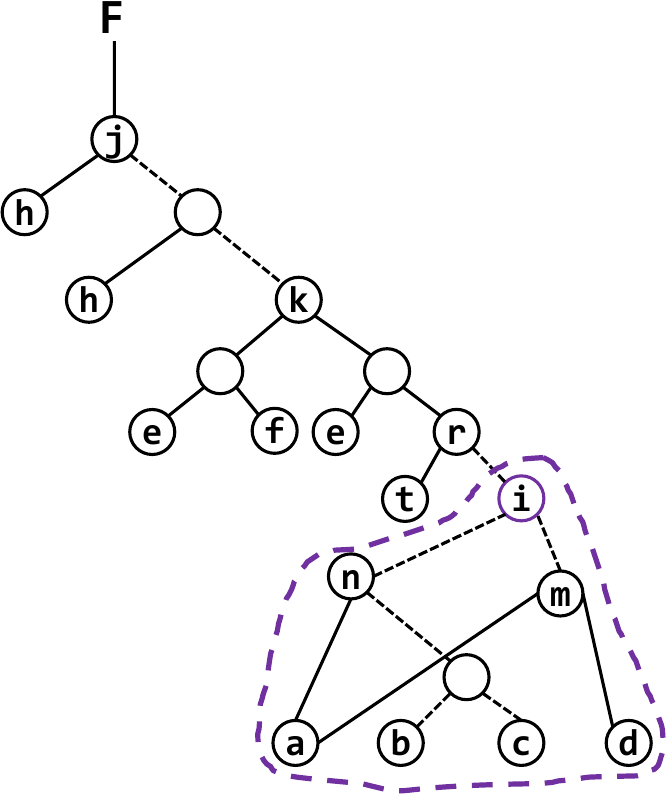}
        \caption{Resubstitution}
        \label{fig:opt_rs}
    \end{subfigure}
    
    \caption{The graph structure augmentation with different optimization methods with the same functionality.}
    \label{fig:optimization}
\end{figure}

\noindent
\textbf{Rewriting}, noted as \texttt{rw}, is a fast greedy algorithm for optimizing the graph size. It iteratively selects the AIG subgraph with the current node as the root node and replaces the selected subgraph with the same functional pre-computed subgraph with a smaller (or equal) size to realize the graph size reduction. Specifically, it finds the 4-feasible cuts as subgraphs for the node while preserving the number of logic levels~\cite{mishchenko2006dag}. As shown in Figure \ref{fig:optimization}, the graph structure alters from Figure \ref{fig:opt_ori} to Figure \ref{fig:opt_rw} when applied with \texttt{rw} at node \texttt{k}.

\noindent
\textbf{Refactoring}, noted as \texttt{rf}, is a variation of the AIG \texttt{rewriting} using a heuristic algorithm~\cite{brayton2006scalable} to produce a large cut for each AIG node. Refactoring optimizes AIGs by replacing the current AIG structure with a factored form of the cut function. It can also optimize the AIGs with the graph depth. The \texttt{rf} optimization applied at node \texttt{j} alters the graph structure as shown in Figure \ref{fig:opt_rf}. 

\noindent
\textbf{Resubstitution}, noted as \texttt{rs}, optimizes the AIG by replacing the function of the node with the other existing nodes (\texttt{divisors}) already present in the graph, which is expected to remove the redundant node in expressing the function of the current node. The optimized graph resulting from \texttt{rs} optimization is shown in Figure \ref{fig:opt_rs}.

Thus, by applying different logic optimization methods to the same AIG, different AIG structures with the same functionality can be generated. In our framework, we further expand the variations by applying different optimizations at each node for combinational augmentation. For example, for an AIG graph with a node size of $N$, the variation space can be $O(3^N)$.

With different graph structures with AIG representation, the netlist variation of hardware design can be realized by technology mapping from various AIGs with the technology library.

\begin{figure*}[!htb]
\centering
    \includegraphics[width=1\textwidth]{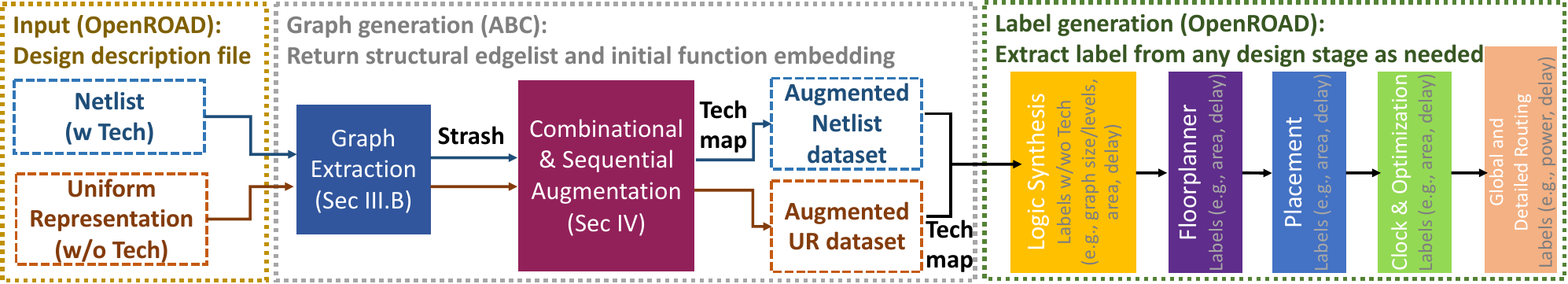}
    \caption{The overview of V2PYG framework. The input files are pre-processed in Yosys as the front-end in OpenROAD. Our framework is implemented in ABC including graph extraction and logic optimization based combinational augmentation, which returns the structural edgelist and initial functional embedding for graph learning. The labels can be extracted from any stage in the downstream OpenROAD design flow.}
    \label{fig:framework_overview}
\end{figure*}

\section{V2PyG Framework}
\label{sec:framework}



\subsection{Overview}
\label{subsec:overview}

The framework overview is shown in Figure \ref{fig:framework_overview}. V2PYG currently performs Verilog to graph representation at the foundation of Boolean representations, including technology dependent (technology mapped) and technology independent representations (Boolean networks such as And-Inv-Graphs). V2PYG parses Verilog designs using Yosys front-end and performs downstream design flow via OpenROAD. First, we represent sequential design as combinational design in a special case. First, transform the sequential design into an equivalent design using Boolean network data structure (e.g., AIGs) or technology dependent networks (exact topology of mapped netlist). Next, model flip-flops as pseudo-primary inputs (PPIs) and pseudo-primary outputs (PPOs), connecting them through the AIG network to capture state transitions and dependencies between flip-flops and combinatorial logic. {The graph is then extracted from the network, and the graph dataset is augmented with combinational methods in logic optimization as shown in Alg. \ref{alg:orch}, which will return the structural information of the edgelist of the graph and the initial functional embedding for graph learning tasks.} Data labels will be generated via OpenROAD \cite{ajayi2019toward} downstream flow. As a result, the sequential design will be represented in semi-complete multi-DAGs. We also plan to integrate MLIR CIRCT \cite{circt} into the pipeline to process multi-fidelity graph representations (DFGs w Boolean networks).

\subsection{Verilog to Graph Representation}
\label{subsec:graph_extract}

\begin{figure*}[!htb]
    \centering
    \begin{subfigure}[b]{0.23\linewidth}
    \centering
        \includegraphics[width=1\textwidth]{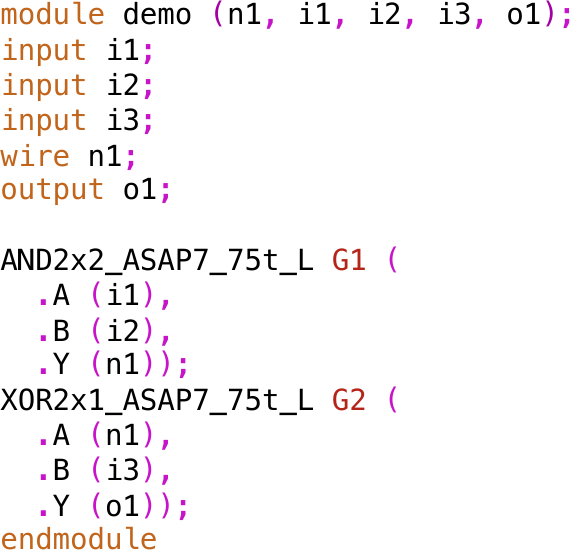}
        \caption{The Verilog described design.}
        \label{fig:ext_net}
    \end{subfigure}
    \hfill
    \begin{subfigure}[b]{0.28\textwidth}
    \centering
        \includegraphics[width=0.7\textwidth]{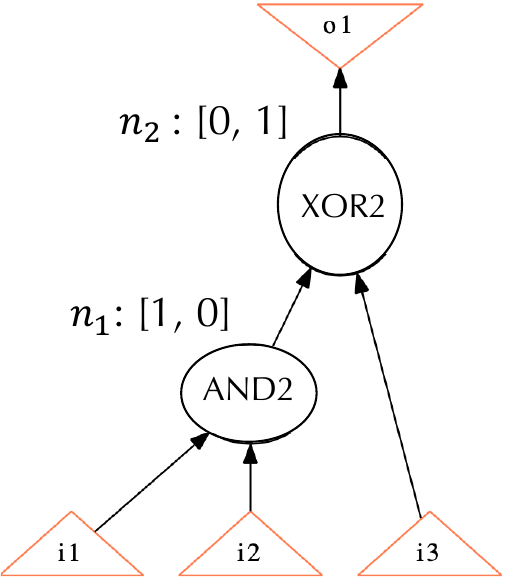}
        \caption{The extracted graph with netlist. Node embedding indicates its gate type.}
        \label{fig:ext_g1}
    \end{subfigure}
    \hfill
    \begin{subfigure}[b]{0.43\textwidth}
    \centering
        \includegraphics[width=0.5\textwidth]{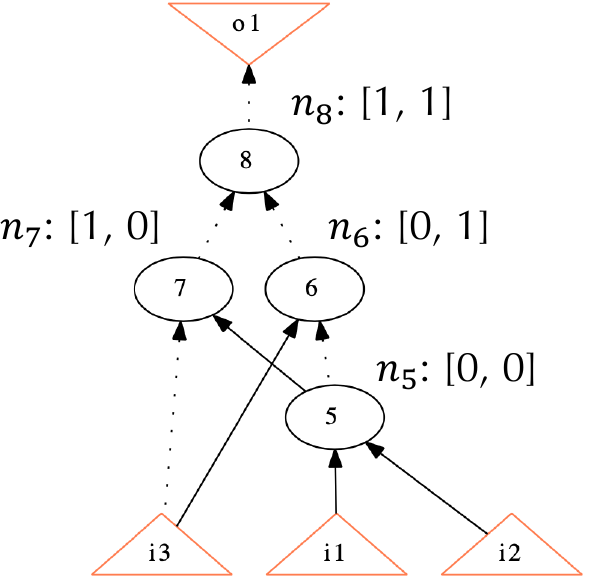}
        \caption{The extracted graph with uniform representation AIG. Node embedding indicates the input inverters to the gate.}
        \label{fig:ext_g2}
    \end{subfigure}
    
    \caption{The extracted graph for GL from different design representation levels. }
    \label{fig:ext_graph}
\end{figure*}

After the prepossessing by Yosys, the design is described at the RTL level, where the gates are connected with nets. As shown in Figure \ref{fig:ext_graph}, in the Verilog file, gate \textit{$G_1$}, which is an \textit{AND2} gate, and gate \textit{$G_2$}, which is an \textit{XOR2} gate, are connected with net \textit{$N_1$}. Thus, we can extract \textit{$G_1$} and \textit{$G_2$} as two nodes, \textit{$n_1$} and \textit{$n_2$}, with features indicting the different gate types. In Figure \ref{fig:ext_g1}, the gate type is encoded with one-hot representation, e.g., $[1, 0]$ indicates the gate type is \textit{AND2} and $[0, 1]$ indicates the gate type is \textit{XOR2} for this design. Net \textit{$N_1$} is formulated as the edge between \textit{n1} and \textit{n2} with the direction from \textit{$n_1$} to \textit{$n_2$} indicating the connection between two gates, i.e., following the topological order, \textit{$n_2$} is at least one level higher than \textit{$n_1$}.

Furthermore, in the logic optimization step, the optimizations are conducted on multi-level technology-independent representations such as And-Inverter-Graphs (AIGs) \cite{mishchenko2006dag,mishchenko2005fraigs, yu2016dag} and Majority-Inverter-Graphs (MIGs) \cite{amaru2015majority, soeken2017exact} of the digital logic, and XOR-rich representations for emerging technologies such as XOR-And-Graphs \cite{ccalik2019multiplicative} and XOR-Majority-Graphs, which means the gate-level netlist is first strashed into uniform representations with the same functionality. For example, to strash the netlist in Figure \ref{fig:ext_net} to AIG, where all gates functions are implemented with only 2-input \textit{AND} gates and inverters, the \textit{XOR2} is strashed into three \textit{AND2} gates and five Inverters. 
Thus, as shown in Figure \ref{fig:ext_g2}, to extract AIGs to graphs, we formulate \textit{AND2} gates as nodes, $n_5$ -- $n_8$, with exact two inputs, and the inverter is encoded at the input net of the \textit{AND2} gate, where the dashed edge indicates an inverter is applied at the net. The initial node embedding indicates the inverters at its input edges from left to right with exact two dimensions with bit representation, where $1$ indicates the inverter at the input and $0$ indicates no inverter at the input. For example, for node $n_8$, there are inverters at both inputs of the gate. Thus, its initial node embedding is $[1, 1]$, while for node $n_7$, there is only one inverter at its left input, the initial node embedding is $[1, 0]$.

For each graph, the framework will return the edgelist for structure information and initial node embedding for the functional information, such as the gate type for the netlist graph and the input inverters for AIGs.

\section{V2PYG RTL Augmentation}

{In the realm of hardware design, specifically at the register-transfer level (RTL), the idea of leveraging machine learning (ML) techniques to enhance optimization processes has garnered increasing attention. However, one significant challenge lies in the acquisition and augmentation of RTL design data to train these ML models effectively. In traditional domains where ML has shown success, such as computer vision or natural language processing, data augmentation techniques—like image rotation, scaling, or text paraphrasing—have been pivotal in enhancing model robustness and generalization. Yet, in the context of RTL design, naive data augmentation can lead to functionally incorrect or non-equivalent designs. Ensuring functional equivalence is paramount; thus, data augmentation in this space requires meticulous care and deep domain knowledge, preventing the straightforward application of techniques that have been successful in other domains. Consequently, while the potential of ML to revolutionize RTL optimization is evident, the unique nature of hardware design data and the stringent requirements for functional correctness substantially complicate the application of traditional data augmentation strategies.}

\subsection{Combinational Augmentation}
\label{subsec:graph_aug}

The combinational augmentation of RTL designs can be achieved through V2PYG by applying transformations that maintain combinational equivalence, inspired by \cite{chowdhury2023almost, chowdhury2023invictus} Specifically, V2PYG incorporates data flow augmentation using polynomial manipulations, fine-grained Boolean algebra manipulation (at least at the individual Boolean node level), and graph isomorphism augmentation. These augmentation techniques enable the generation of a nearly unlimited number of functionally equivalent designs at the Boolean level.
As discussed in Section \ref{sec:background}, applying different logic optimizations to the same functional AIG can result in various graph structures. However, stand-alone logic optimizations are limited in graph augmentation. With three optimizations, i.e., \textit{rw}, \textit{rs}, \textit{rf}, only three variations are generated for the same original AIG. Thus, we propose combinational augmentation, where we randomly apply different optimizations at each AIG node for graph structure alterations. For an original AIG with a node size of $N$, the potential graph variations can be at the scale of $O(3^N)$, and the variation space increases exponentially as the graph size increases.

The algorithm for augmentation with combinational logic optimization methods is shown in Alg. \ref{alg:orch}. It takes the original AIG $G(V, E)$ and a random seed as inputs. Following the topological order in the original graph, it first initializes a list $D$ for recording the available optimizations, where we hash the optimizations with numbers $0$, $1$, $2$, $3$, i.e., $0$ indicates none of the optimizations is applied, which is initialized for all AIG nodes to be considered in the random pool; $1$ indicates \textit{rw} is applicable for logic optimization at the node; $2$ indicates \textit{rf} is applicable for optimization at the node; $3$ indicates \textit{rs} is applicable for optimization at the node. Then, for each AIG node, it generates the list $D$ for random selections by checking the transformability for each optimization operation from line 3 to line 8. After acquiring the list $D$ with all available optimizations, it randomly selects the optimization from $D$ (line 9) and updates the graph accordingly (line 10). Note that the resulting AIG structures from different optimizations are checked to be functional equivalence by Combinatorial Equivalence Checking (CEC).

\begin{algorithm}
\caption{Boolean manipulation sampling on AIGs}   
\label{alg:orch}
 
   \SetKwInOut{KwIn}{Input}
     \SetKwInOut{KwIn}{Input}
    \SetKwInOut{KwOut}{Output}
    \KwIn{~$G(V,E) \leftarrow$ Boolean Networks/Circuits in AIG} 
    \KwIn{Random seed for optimization selections}
    \KwOut{~Post-optimized AIG $G(V,E)$}
 \For{$v \in V$ in topological order}{
     $D$ = [0]  \tcp{List $D$ saves the available optimization methods at the node. The length of $D$ can vary.}
  \uIf{v \text{is transformable w.r.t} $rw$}{
        D.append(1) \tcp{$rw$ is available for optimization.}}
  \uIf{v \text{is transformable w.r.t} $rs$}{
        D.append(2) \tcp{$rf$ is available for optimization.}}
  \uIf{v \text{is transformable w.r.t} $rf$}{
        D.append(3) \tcp{$rs$ is available for optimization.}}
        int k = rand(0, len($D$)) \tcp{Randomly select an optimization from $D$.}

  Update $G(V,E) \xleftarrow{D[k]}$ \texttt{Dec\_GraphUpdateNetwork} with the selected optimization $D[k]$, and exclude $v$ and transformed nodes from $V$.
  }
\end{algorithm}

\begin{figure*}[!htb]
    \centering
    \begin{subfigure}[b]{0.3\linewidth}
    \centering
        \includegraphics[width=0.5\textwidth]{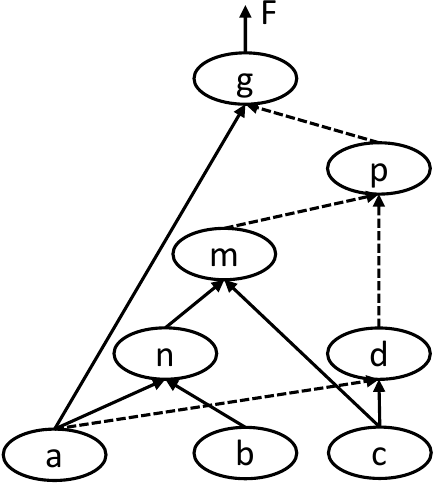}
        \caption{The original AIG.}
        \label{fig:aug_g1}
    \end{subfigure}
    \hfill
    \begin{subfigure}[b]{0.3\textwidth}
    \centering
        \includegraphics[width=0.8\textwidth]{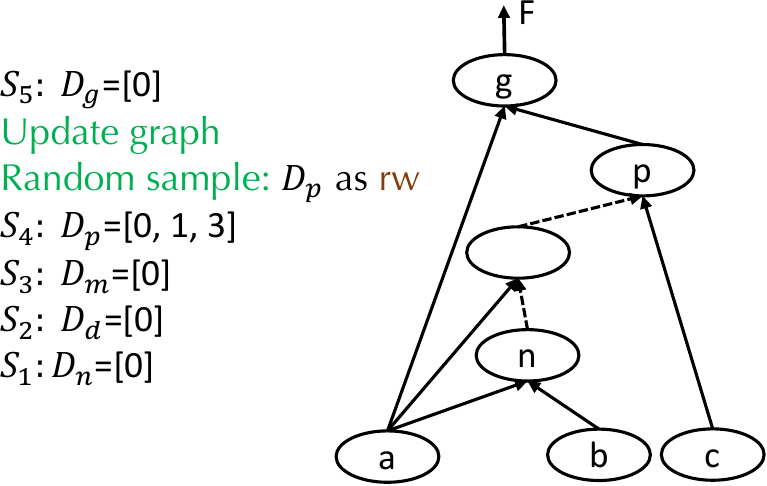}
        \caption{The first graph variation with combinational augmentation.}
        \label{fig:aug_g2}
    \end{subfigure}
    \hfill
    \begin{subfigure}[b]{0.3\textwidth}
    \centering
        \includegraphics[width=0.9\textwidth]{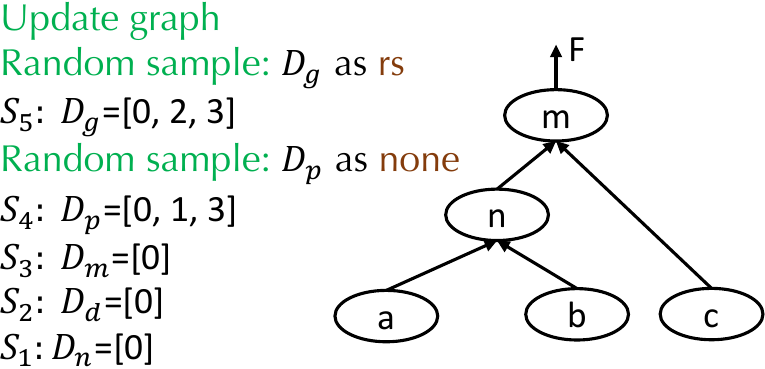}
        \caption{The other graph variation with combinational augmentation.}
        \label{fig:aug_g3}
    \end{subfigure}
    
    \caption{The combinational augmentation examples. }
    \label{fig:aug_graph}
\end{figure*}

For example, in Figure \ref{fig:aug_graph}, the original AIG is shown in Figure \ref{fig:aug_g1}. The algorithm checks the optimization transformability following the topological order starting from $S_1$ as shown in Figure \ref{fig:aug_g1}. For node $n -- m$, they have no applicable optimizations for selection and are skipped for optimization. Regarding node $p$, it poses three optimizations, and the random selection can result in quite different graph structures for dataset augmentation. In the first random sample, the algorithm picks $rw$ for the optimization at node $p$ and the graph is updated to Figure \ref{fig:aug_g2} thus, node $g$ shows no applicable optimizations. In the second random sample, the algorithm picks $none$ for the optimization at node $p$, resulting in optimization opportunities for node $g$. When the algorithm picks $rs$ for the optimization at $g$, the graph is updated as shown in Figure \ref{fig:aug_g3}. By setting different random seeds and starting the random sample with a push button implemented in our framework, the dataset is augmented through combinational random selections with functional equivalence.

\begin{figure}[!htb]
    \centering
    \begin{subfigure}[b]{0.24\textwidth}
    \centering
        \includegraphics[width=1\textwidth]{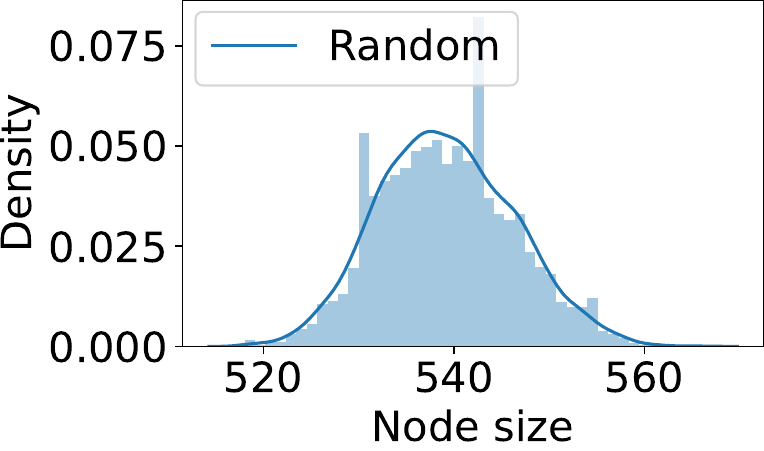}
        \caption{b11}
        \label{fig:sample_b11}
    \end{subfigure}
    \hfill
    \begin{subfigure}[b]{0.24\textwidth}
    \centering
        \includegraphics[width=1\textwidth]{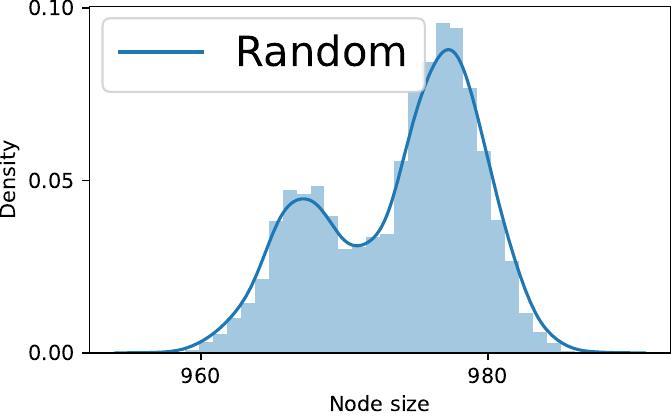}
        \caption{b12}
        \label{fig:sample_b12}
    \end{subfigure}
    \hfill
    \begin{subfigure}[b]{0.24\textwidth}
    \centering
        \includegraphics[width=1\textwidth]{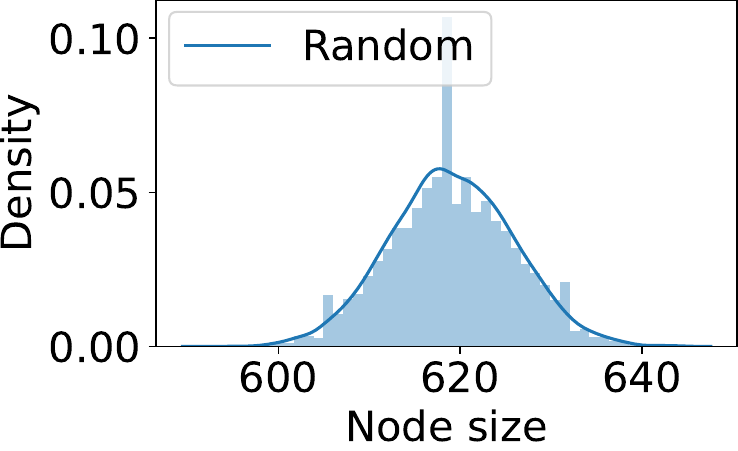}
        \caption{c2670}
        \label{fig:sample_c2670}
    \end{subfigure}
    \hfill
    \begin{subfigure}[b]{0.24\textwidth}
    \centering
        \includegraphics[width=1\textwidth]{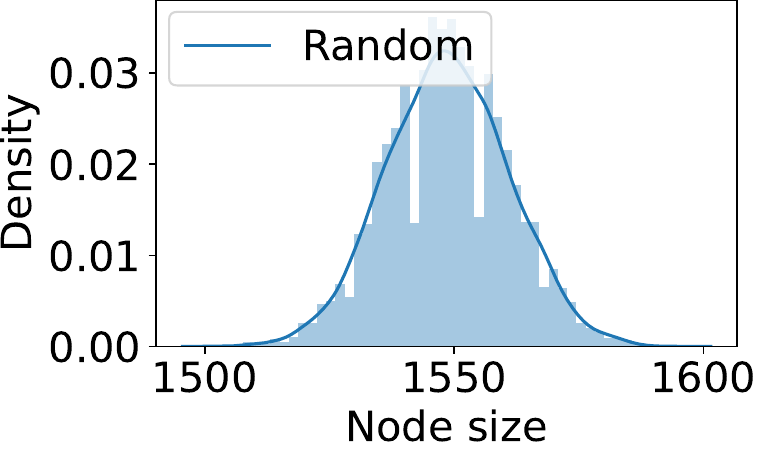}
        \caption{c5315}
        \label{fig:sample_c5315}
    \end{subfigure}
    
    \caption{The optimization quality distribution with 6000 samples of purely random sampling and priority guided sampling. 
    }
    \label{fig:sample}
\end{figure}

Furthermore, we show the dataset distribution with 6000 samples in Figure \ref{fig:sample} for different designs with Alg. \ref{alg:orch}. Here, we collect the distribution w.r.t the graph size of various graph structures, which is reported by the logic synthesis in OpenROAD as labels. By combinational dataset augmentation, we can generate large enough datasets, and with the proposed random selected logic-optimization-based boolean manipulations on graphs, we can generate Gaussian-like dataset distributions for effective GL model training efficiently.


\subsection{Sequential Augmentation} 

V2PYG currently supports sequential augmentation by adjusting the sequential behaviors using partial random retiming, which offers significant design variations while preserving sequential equivalence. It is important to note that combinational augmentation is orthogonal to sequential augmentation. 

In the sequential augmentation, we use the variations in \texttt{retiming}, where we first check the feasibility of each edge for buffer insertion and randomly insert the buffer at the applicable edge, i.e., resulting in random partitions of the sequential graph. After sequential augmentation, we can further apply combinational augmentation on each sub-graph resulting from sequential augmentations.

\subsection{Label Generation}

{
In Figure \ref{fig:sample}, multiple labels can be derived for a specific RTL design. For example, the quality metrics related to the AIG size, demonstrated in Figure \ref{fig:sample}, signify one category of labels formulated during the logic synthesis stage of ABC. Alongside technology-independent representation labels, technology-dependent labels at the logic level are adeptly generated through design infrastructures such as ABC. Other ABC-compatible platforms like OpenROAD \cite{ajayi2019openroad}, Yosys \cite{wolf2016yosys}, and VTR \cite{murray2020vtr} also support this functionality. By harnessing technology mapping in tools like ABC and Yosys and utilizing methods such as LUT-mapping (e.g., \texttt{'if'} in ABC) or standard-cell mapping (e.g., \texttt{'map'}), we can generate labels for metrics encompassing the number of LUTs, LUT-based netlist depth, and delay and area metrics of the standard-cell netlist.

Additionally, while V2PYG is inherently integrated with ABC, it offers potential integration with OpenROAD to acquire downstream data labels, including those from floorplanning, placement, and routing stages. Specifically, via the Yosys ABC interface (command \texttt{abc} in Yosys), V2PYG functionalities can be seamlessly incorporated into the OpenROAD flow. In this scenario, we can garner custom labels at specific design phases, virtually spanning the complete design procedure. For instance, each sampled AIG graph can be channeled to OpenROAD for subsequent design phases like technology mapping, floorplanning, placement, and routing. Following each stage, OpenROAD renders a design report, which can be mined to extract labels, thereby training GNN models tailored for diverse tasks.}

\section{V2PYG Implementation}
\label{sec:implementation}

\subsection{Implementation}
\label{subsec:abc_imp}

\paragraph{Graph Extraction} We have introduced a universal extraction implementation \texttt{write\_edgelist}, which supports the extraction for AIGs, standard-cell techmap netlist, and LUT-based netlist. More details of the usage can be found by \texttt{write\_edgelist -h} in ABC and the following user case examples.

\paragraph{Logic-level augmentation} This framework has been implemented in ABC~\cite{brayton2010abc} with the command \texttt{aigaug}, which conducts logic optimization based combinational augmentation of the design with uniform representations following Alg. \ref{alg:orch}. Specifically, we implement five flags in \texttt{aigaug} -- (1) `\texttt{-s}` takes in the customized random seed for the random selections; (2) `\texttt{-d}` sets the name of the file recording the random selected and applied optimization operation at each node; (2) `\texttt{-z}` enables the zero usage for \texttt{rewrite} optimization, where the graph structure is altered even when there is no graph size reduction by \texttt{rewrite}; (4) `\texttt{-Z}` enables the zero usage for \texttt{refactor} optimization, where the graph structure is altered even when there is no graph size reduction by \texttt{refactor}; (5) `\texttt{-help}` prints out the usage of the command.

\subsection{User Example 1 - Combinational RTL Designs}
\label{subsec:use_exp1}

In this section, we present the toy example that demonstrates a series of functions associated with V2PYG implemented in ABC using a 2-bit multiplier. Our starting point is an RTL implementation of a 2-bit multiplier shown in Listing \ref{lst:2bitMultiplierv}. 

\begin{lstlisting}[caption={RTL description of a 2-bit multiplier.},label=lst:2bitMultiplierv]
module mult2b (a,b,z);
  input [1:0] a,b;
  output [3:0] z;
    assign z = a * b;
endmodule
\end{lstlisting}

To translate the word-level description of our 2-bit multiplier into an AIG representation, we employ a sequential procedure within the ABC tool, as illustrated in Listing \ref{lst:ABCCommands}.

\begin{lstlisting}[caption={Procedure in ABC for transforming the RTL design into an AIG.},label=lst:ABCCommands]
abc 01> %read mult-2b.v ; %blast; &put; strash; print_stats;
mult2b     : i/o =    4/    4  lat =    0  and =     10  lev =  4
abc 02> write mult-2b.blif
\end{lstlisting}

In the above commands, `\texttt{\%read mult-2b.v ; \%blast}` facilitates the reading of our multiplier's Verilog code, subsequently bit-blasting it to a bit-level representation. The sequence `\texttt{\&put;strash}` then morphs the design into its AIG avatar. Lastly, the user can serialize the AIG representation to a BLIF format using `\texttt{write mult-2b.blif}`. This pedagogical demonstration serves as a testament to ABC's prowess in handling and transforming RTL designs, marking its indelible significance in the RTL design landscape.


\paragraph{AIG} 

\begin{lstlisting}[caption={Procedure of extracting AIG representation.},captionpos=b]
./abc
abc 01> read mult-2b.blif
abc 02> strash
abc 03> write_edgelist mult-2b.el
WriteEdgelist (Verilog-to-PyG @ https://github.com/ycunxi/Verilog-to-PyG) starts writing to mult-2b.el
abc 04> write_edgelist -h
usage: write_edgelist [-N] <file>
             writes the network into edgelist file
             part of Verilog-2-PyG (PyTorch Geometric). more details https://github.com/ycunxi/Verilog-to-PyG
    -N     : toggle keeping original naming of the netlist in edgelist (default=False)
    -h     : print the help massage
    file   : the name of the file to write (extension .el)
\end{lstlisting}


\begin{multicols}{2}
\begin{lstlisting}[numbers=left, stepnumber=1, numberfirstline=false, label=lst:aig-edgelist]
1 9 Pi 00
2 10 Pi 00
3 11 Pi 00
4 12 Pi 00
9 23 AIG 11
11 23 AIG 11
10 27 AIG 11
11 27 AIG 11
9 28 AIG 11
12 28 AIG 11
27 29 AIG 11
28 29 AIG 11
27 30 AIG 00
28 30 AIG 00
\end{lstlisting}

\begin{lstlisting}[numbers=left, stepnumber=1, numberfirstline=false, firstnumber=15]
29 24 AIG 00
30 24 AIG 00
10 31 AIG 11
12 31 AIG 11
29 26 AIG 11
31 26 AIG 11
29 32 AIG 00
31 32 AIG 00
26 25 AIG 00
32 25 AIG 00
23 5 Po 00
24 6 Po 00
25 7 Po 00
26 8 Po 00
\end{lstlisting}
\end{multicols}

\begin{figure}[!htb]
    \centering
        \includegraphics[width=0.45\textwidth]{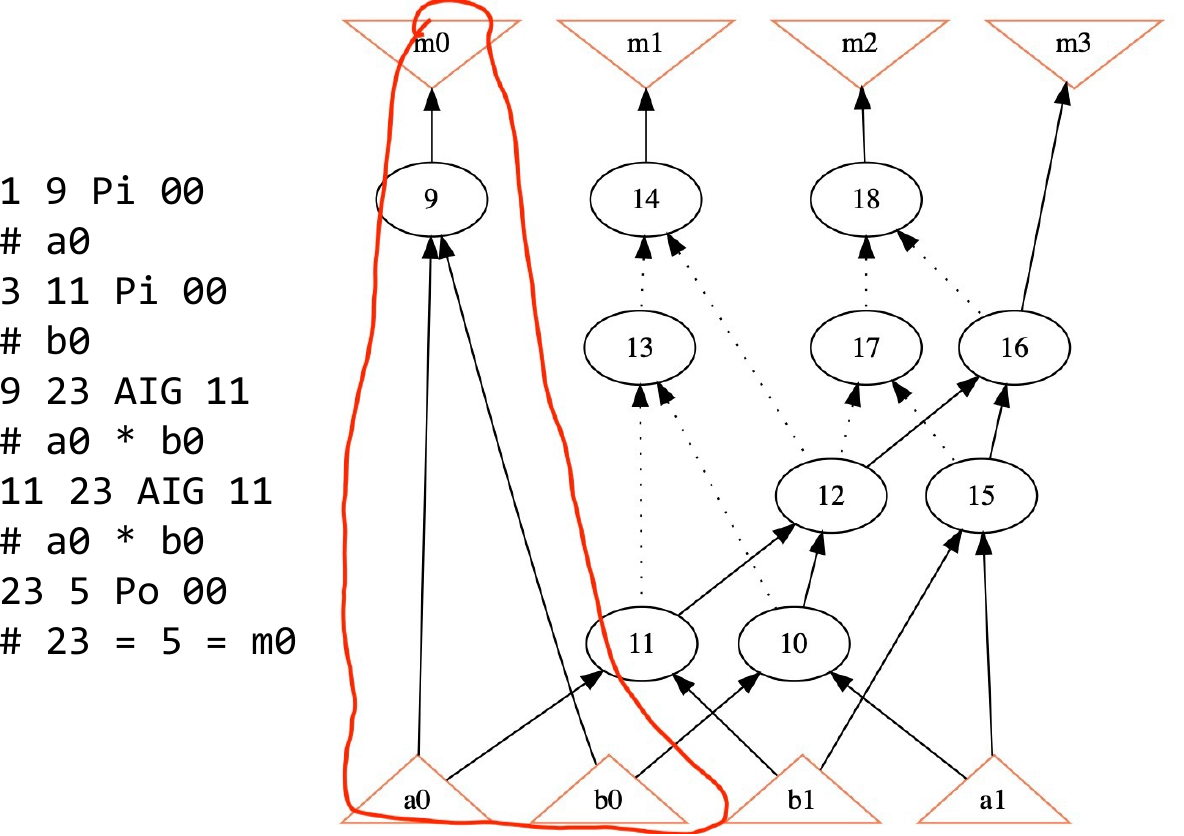}
        \caption{Visualization of the topological representation and static AIG features w.r.t Listing \ref{lst:aig-edgelist}.}
        \label{fig:2-b-mult-aig-trace}
\end{figure}

\paragraph{Gate-level netlist}

\begin{figure*}[!htb]
    \centering
        \includegraphics[width=0.98\textwidth]{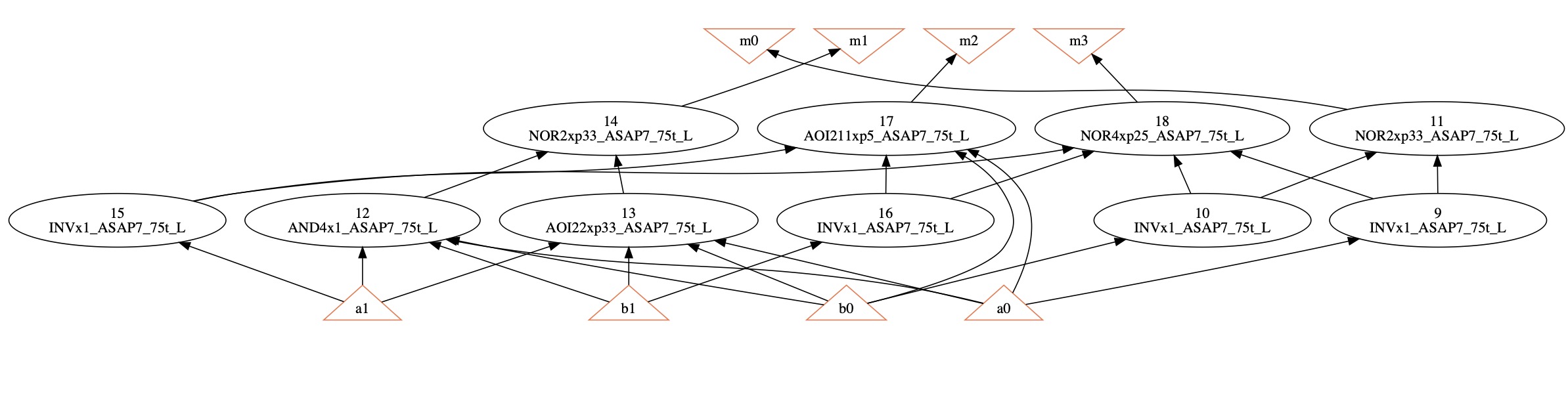}
        \caption{Visualization of the topological representation of technology mapped netlist (ASAP7 library \cite{xu2017standard} used in this example).}
        \label{fig:2-b-mult-mapped-trace}
\end{figure*}

Next, we will illustrate a case study in ABC for generating an edge list of the mapped netlist. This list succinctly encapsulates the interconnections of the design's elements, providing a blueprint of its topology. The format and contents of the edge list for our 2-bit multiplier are showcased in Listing \ref{lst:edgeList}, which can be matched from Figure \ref{fig:2-b-mult-mapped-trace}. In addition to the topological information (connectivity), the edgelist contains the exact cell entry of the technology library. In this example, the cell matches the ASAP7 library, e.g., INVx1\_ASAP7\_75t\_L indicates the functionality of the cell, the size of the cell, and its library information. Note that we plan to further expand the functionality of feature extract for technology mapped cells, such as adding more feature columns for the cell information, such as capacitor size, rise/fall timing, area, etc.  

\begin{lstlisting}[caption={Verilog of technology-mapped netlist.},label=lst:edgeList]
module Multi2 ( 
    a0, a1, b0, b1,
    m0, m1, m2, m3  );
  input  a0, a1, b0, b1;
  output m0, m1, m2, m3;
  wire new_n9_, new_n10_, new_n12_, new_n13_, new_n15_, new_n16_;
  INVx1_ASAP7_75t_L         g0(.A(a0), .Y(new_n9_));
  INVx1_ASAP7_75t_L         g1(.A(b0), .Y(new_n10_));
  NOR2xp33_ASAP7_75t_L      g2(.A(new_n9_), .B(new_n10_), .Y(m0));
  AND4x1_ASAP7_75t_L        g3(.A(b0), .B(a0), .C(a1), .D(b1), .Y(new_n12_));
  AOI22xp33_ASAP7_75t_L     g4(.A1(a0), .A2(b1), .B1(a1), .B2(b0), .Y(new_n13_));
  NOR2xp33_ASAP7_75t_L      g5(.A(new_n13_), .B(new_n12_), .Y(m1));
  INVx1_ASAP7_75t_L         g6(.A(a1), .Y(new_n15_));
  INVx1_ASAP7_75t_L         g7(.A(b1), .Y(new_n16_));
  AOI211xp5_ASAP7_75t_L     g8(.A1(b0), .A2(a0), .B(new_n15_), .C(new_n16_), .Y(m2));
  NOR4xp25_ASAP7_75t_L      g9(.A(new_n15_), .B(new_n9_), .C(new_n16_), .D(new_n10_), .Y(m3));
endmodule
\end{lstlisting}

\begin{lstlisting}[caption={Procedure in generating edgelist of technology mapped netlist},captionpos=b]
./abc
abc 01> read 7nm_lvt_ff.lib
Library "ASAP7_7nm_LVT_FF" from "7nm_lvt_ff.lib" has 159 cells (26 skipped: 23 seq; 0 tri-state; 3 no func; 0 dont_use).  Time =     0.70 sec
Warning: Detected 2 multi-output gates (for example, "FAx1_ASAP7_75t_L").
abc 01> read -m mult-2b-mapped.v
abc 02> write_edgelist mult-2b-mapped.el
\end{lstlisting}

\begin{lstlisting}[caption={Edgelist of technology mapped netlist.},captionpos=b]
1 9 Pi 00
2 10 Pi 00
3 11 Pi 00
4 12 Pi 00
9 27 INVx1_ASAP7_75t_L
11 28 INVx1_ASAP7_75t_L
27 28 23 NOR2xp33_ASAP7_75t_L
11 9 10 12 29 AND4x1_ASAP7_75t_L
9 12 10 11 30 AOI22xp33_ASAP7_75t_L
30 29 24 NOR2xp33_ASAP7_75t_L
10 31 INVx1_ASAP7_75t_L
12 32 INVx1_ASAP7_75t_L
11 9 31 32 25 AOI211xp5_ASAP7_75t_L
31 27 32 28 26 NOR4xp25_ASAP7_75t_L
23 5 Po 00
24 6 Po 00
25 7 Po 00
26 8 Po 00
\end{lstlisting}




\subsection{Equivalence Verification of Augmented Designs}

{In the intricate domain of RTL design, demonstrating functional equivalence in data-augmented designs is paramount. To elucidate this, consider the scenario executed using the \texttt{abc} tool, renowned for And-Inverter Graph (AIG) manipulation. Initially, the design \texttt{i10.aig} is loaded. Post structural hashing (\texttt{strash}), the design encapsulates 257 inputs, 224 outputs, and leverages 2675 AND gates. To perform data augmentation, a random seed of `0` is deployed, birthing the augmented design \texttt{i10\_aug\_0.aig}. The statistics of this new design reveal modifications, notably a decrement in AND gates to 2514, albeit retaining an identical level depth. Mirroring this procedure but employing a divergent random seed of `1`, we derive another offspring, the \texttt{i10\_aug\_1.aig}. The design, now slightly deviant, holds 1980 AND gates within a level depth of 47. Most crucially, upon deploying the combinational equivalence checking (\texttt{cec}), results affirm that the parent \texttt{i10.aig} and its progenies \texttt{i10\_aug\_0.aig} and \texttt{i10\_aug\_1.aig} are, in essence, functionally equivalent. This not only underscores the power and precision of such tools but also the intricate art of RTL data augmentation where functional integrity remains inviolate.}

\begin{lstlisting}
//load i10.aig design
abc 01> read i10.aig;strash;print_stats;
i10: i/o =  257/  224  lat =    0  and =   2675  lev = 50
//perform augmentation using random seed "0"
abc 03> aigaug -s 0 -d i10_aug_0.csv;print_stats
i10: i/o =  257/  224  lat =    0  and =   2514  lev = 50
abc 03> write i10_aug_0.aig;
abc 03> read i10.aig;strash;
//perform augmentation using random seed "1"
abc 05> aigaug -s 1 -d i10_aug_1.csv;print_stats
i10: i/o =  257/  224  lat =    0  and =   1980  lev = 47
abc 05> write i10_aug_1.aig;
abc 05> cec i10.aig i10_aug_0.aig;
Networks are equivalent.  Time =     0.19 sec
abc 05> cec i10.aig i10_aug_1.aig;
Networks are equivalent.  Time =     0.24 sec
//combinational equivalence checking returns that all three designs are equivalent
\end{lstlisting}

\section{Conclusion}

This paper proposes a novel framework V2PYG to address the growing complexities in modern hardware designs by introducing advanced methodologies for optimizing and analyzing digital systems. We have introduced an innovative, open-source framework known as Verilog-to-PyG (V2PYG) that enables seamless integration with the PyTorch Geometric graph learning platform \cite{fey2019fast}. This framework also offers compatibility with the open-source Electronic Design Automation (EDA) toolchain OpenROAD, Yosys, and VTR, thereby facilitating the collection of labeled datasets in a fully open-source environment. Moreover, we presented novel techniques for RTL data augmentation that can be incorporated into our framework. These methods contribute to the development of an extensive database of graph-based RTL designs validated in equivalence checking verification.

\bibliographystyle{IEEEtranS}
\bibliography{bib/Yingjie.bib, refs.bib}

\end{document}